# Self-attention based BiLSTM-CNN classifier for the prediction of ischemic and non-ischemic cardiomyopathy

Kavita Dubey[1], Anant Agarwal[2], Astitwa Sarthak Lathe[1], Ranjeet Kumar[3] and Vishal Srivastava[1]*

1 Electrical and Instrumentation Engineering Department, Thapar Institute of Engineering and Technology, Patiala, India
2 Electronics and Communication Engineering Department, Thapar Institute of Engineering and Technology, Patiala, India
3 University of St Andrews, Scotland

*Corresponding author: vishalsrivastava17@gmail.com

**Abstract:** Heart Failure is a major component of healthcare expenditure and a leading cause of mortality worldwide. Despite higher inter-rater variability, endomyocardial biopsy (EMB) is still regarded as the standard technique, used to identify the cause (e.g. ischemic or non-ischemic cardiomyopathy, coronary artery disease, myocardial infarction etc.) of unexplained heart failure. In this paper, we focus on identifying cardiomyopathy as ischemic or non-ischemic. For this, we propose and implement a new unified architecture comprising CNN (inception-V3 model) and bidirectional LSTM (BiLSTM) with self-attention mechanism to predict the ischemic or non-ischemic to classify cardiomyopathy using histopathological images. The proposed model is based on self-attention that implicitly focuses on the information outputted from the hidden layers of BiLSTM. Through our results we demonstrate that this framework carries a high learning capacity and is able to improve the classification performance.

***Keywords*:** Convolutional Neural Network (CNN), Inception-V3, Long Short-Term Memory (LSTM), Self-Attention, Histopathological Images and Cardiomyopathy.

## Introduction

Cardiovascular disease (CD) is a major cause of heart failures and related causalities worldwide. According to the survey done by Centres for Disease Control (CDC) in 2011, CD is the foremost cause of deaths in the United States, Australia, United Kingdom and Canada [1]. Heart failure is a severe, progressive clinical syndrome that results in inadequate systemic perfusion. It is evident through other common symptoms like arrhythmia (irregular heartbeats), myocardial infarction (commonly known as heart-attack), and coronary heart disease [2].

Cardiomyopathy is responsible for approximately one third of all clinical heart failure cases. It is a disease related to hardening of the heart muscle that leads to heart failure in several cases [3]. Depending upon the underlying cause, cardiomyopathy has been further divided into two categories: ischemic and non-ischemic cardiomyopathy [4]. Ischemic and idiopathic dilated cardiomyopathy (a type of non-ischemic cardiomyopathy) are the two most common causes of heart failure with left ventricular systolic dysfunction, the only definite treatment being the heart transplant. Ischemic cardiomyopathy is a coronary heart disease caused by left ventricle dysfunction due to chronic absence of oxygen to the heart muscle. On the other hand, non-ischemic cardiomyopathy is not caused by coronary artery disease and is often associated with organ illnesses and exhibits common symptoms such as; breathlessness, sweating, fast breathing, high level of fatigue etc. [5]. The gold standard method for the differentiation between ischemic and non-ischemic cardiomyopathy is coronary angiography [6]. However, due to its high cost and invasive nature, it is not feasible to analyze all patients with systolic heart failure by coronary angiogram. In order to prevent unnecessary coronary angiography, it would be very useful to be able to differentiate patients suffering from non-ischemic cardiomyopathy non-invasively with adequate precision. Non-invasive techniques such as thallium scintigraphy are costly, while dobutamine-stressed echocardiography is technician-dependent and is not accessible in general [6-8]. The other non-invasive imaging modalities like compute tomography (CT), magnetic resonance imaging (MRI) are preferred choices because of improved image quality and diagnostic accuracy [7-12]. In addition to CT and MRI, an endomyocardial biopsy (EMB) is considered in some instances complementary; or is the only procedure for diagnosing both ischemic and non-ischemic conditions in unexplained cardiomyopathy cases [13-15]. The EMB based diagnostics, depends upon the conviction of pathologists and hence there is an ample room for inter-rater variability [16]. The quantitative interpretation of pathology images is very important for an accurate diagnosis [17]. At present, quantitative analysis in a standardized clinical environment is a time-consuming affair and carries high inter-rater variability, and so an automated quantification algorithm fostering diagnostic outcomes with universal acceptability is urgently required.

Recent, burgeoning and successful applications of deep learning in the field of medicine and digital pathology demonstrate its effectiveness in learning hidden patterns that may not be visible by human examination [16]. In fact, deep learning architectures have been shown to be successful in the automated classification and segmentation of disease in digital histopathology [17]. The benefit of models based on deep learning is that they become familiar with the most appropriate representation progressively, as part of the training process. However, in the traditional CNN's output layer is fully connected to the hidden layer, but it includes inefficient or multiple kernel that extract the same or trivial information repeatedly from input data [17]. To improve the accuracy further, network can be enlarged by adding more kernels, convolutional layers, and pooling layers, but it will increase the computational cost and also there is a chance of overfitting [18, 19]. LSTM is more appropriate for handling time-dependence problems [20]. It can filter and fuse the empty input, similar information, and irrelevant information extracted from the convolutional kernels, so that the effective and relevant information can be stored for a long time in the state cell. Benefit of a memory cell and controlling gates is that it prevents the gradient from vanishing too quickly, thereby propagating information without loss [21-23]. But LSTM only exploits the preceding or past information. BiLSTM is an advancement of LSTM in which forward hidden layer is combined with backward hidden layer, that can access both the preceding and subsequent information. The vector representation of high intra-variability and sparsity causes histopathological images of high-dimensional vector [24]. The high-dimensional vector acting as the BiLSTM input will raise the number of network parameters thus making it harder to optimize the network. The convolution operation reduces the dimensionality of the data [25-27]. Although, BiLSTM with CNN can obtain the contextual information, but it can't focus on the relevant information extracted from contextual information [28]. The attention mechanism by setting the distinct weights can highlight the important information from the acquired information. The mechanism of attention is an efficient method which enables a model to pay more attention to significant information. The integration of BiLSTM-CNN with self-attention mechanism can further enhance the performance of the model.

In this manuscript, we report a novel framework based on Inception-V3 (an example of CNN) and BiLSTM with self-attention for the prediction of ischemic and non-ischemic cardiomyopathy. The proposed framework has an intrinsic self-attention ability, i.e., using the feature maps acquired from high-layers as the attention mask of a lower layers, rather than learning the attention mask with additional layers. Results reveal that the proposed model attain high performance with histopathology images than the conventional CNN approach.

**Methodology**

**Preparation of Input Image**: Images of the size 96 x 96 pixels are considered as an input image to the network. High-resolution histopathological images which can be easily annotated by the pathologist are used to generate a binary mask. The approach is associated with binary image categorization, for each case features like textural properties, colour or spatial features. In our binary mask, '1' is assigned to ischemic cardiomyopathy and '0' is assigned to non-ischemic cardiomyopathy. The dataset was split randomly, i.e., 65 images were used for training, and 29 images were held out as an independent test set.

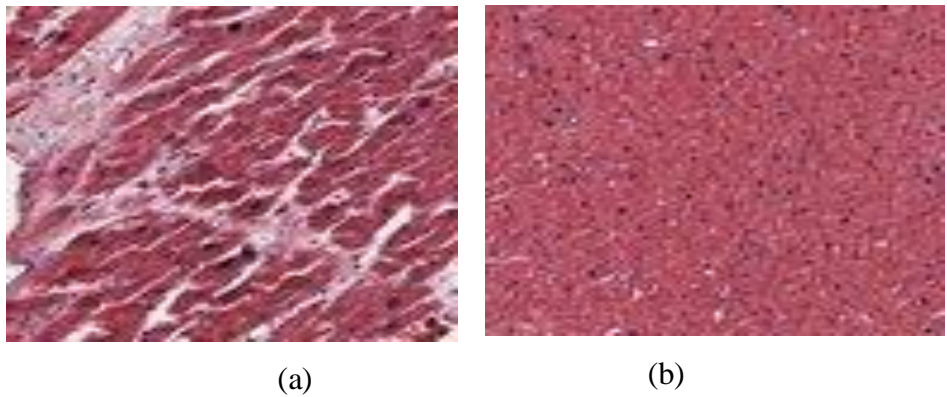

(a)  (b)

**Fig. 1** (a) Ischemic and (b) Non-ischemic histopathology images of heart.

**Model Description:** The structure of the model presented in this manuscript was implemented in KERAS. The architecture of the model consists of Inception-V3, BiLSTM cells and self-attention mechanism, and FCNN as shown in the Fig. 2. Input to the inception-V3 module is an image of the size 96 x 96 pixels. Inception-V3 (an example of CNN) consists of convolutional layers that extract features from the input images. The goal of the inception module is to work as "multi-level feature extractor" by computing 1×1, 3×3, and 5×5 convolutions within the same module of the network. A reduction in the number of features involved without the loss of information allows CNN to generate an appropriate representation of the dataset. Along the channel dimensions, the output of these filters is stacked before being fed into the next layer of the network. To avoid the problem of overfitting, convolutional layers are followed by dropout layers during training with dropping probability of 0.5 and Kernel regularization is done by using L2 regularization with λ= 0.01 [24]. Further, data augmentation is used to enlarge the size of datasets as well as to ameliorate the overfitting issue [25]. We perform augmentation by flipping, rotation and zoom scaling without altering the actual features of the images. Padding is done to ensure the input and output are of the same size. *Tanh* function is used as activation functions in all the layers of the inception-V3 network. Adam optimizer with a learning rate of 0.0001 and the decay-rate of 0.000001 is used as an optimizing function for the proposed model. The model is trained until convergence is achieved. Reshaping process is required before giving the output of CNN as an input to the BiLSTM. Inception-V3 have 2-D representation at its output that is reshaped to 1D array before passing the output to BiLSTM, which accepts 1D in its input. The memory cell of BiLSTM stores the feature information extracted from inception-V3 during current and previous training iterations that is accessed by different controlling gates (like input gate, output gate, forget gate) [26-28]. In the proposed model, the input to the BiLSTM is a 1D array of size 1 x 2048. By knowing the input, past, future states of its local neighbours, Bi-LSTM can predict the present state. Dropout of 0.5 is applied on the Bi-LSTM layer (input, output as well as on the image representation vector).

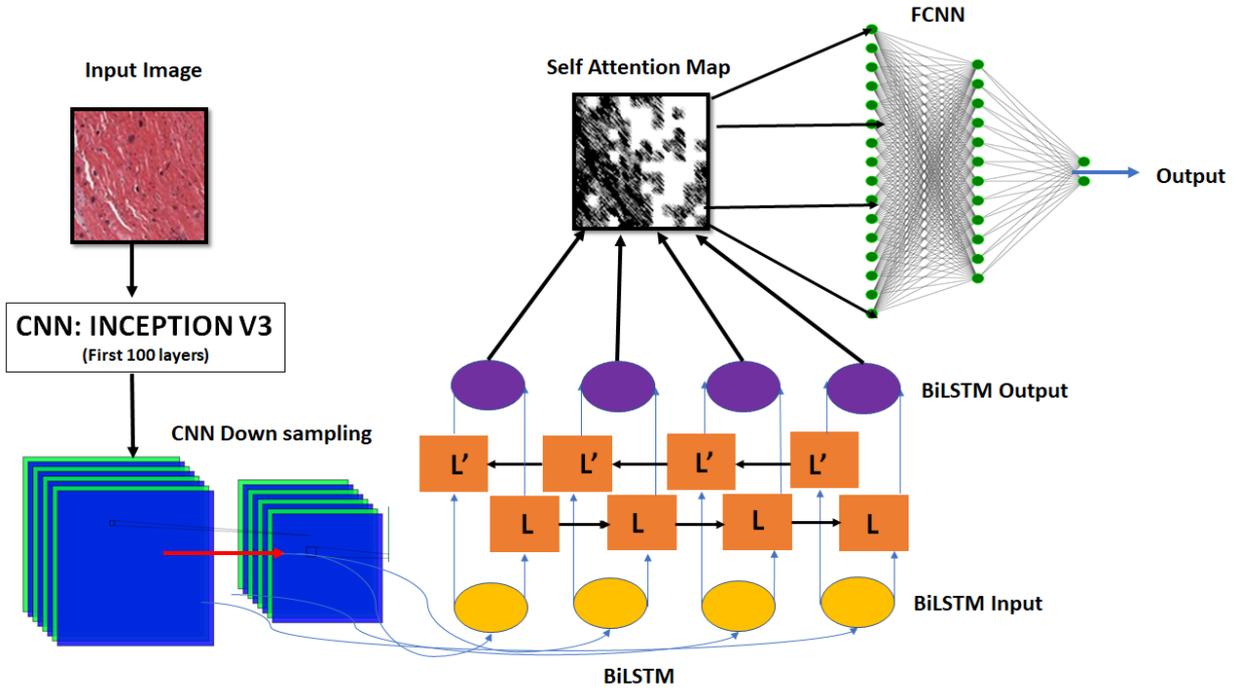

**Fig.2.** Architecture of the proposed model, L & L': Convolutional Layers.

A process called self-attention mechanism is performed on the output of BiLSTM to obtain self-attention map. It executes 'attention process' at every position of the input sequence. The attention process is carried out by computing a context vector for a decoder that contains the most useful information from all hidden states of the encoder, with an averaging of weights done on. A self-attention model with a 1-layer stacked configuration has been implemented in the network structure. It works with the aim to find an effective classification in terms of accuracy, sensitivity and specificity. Query, key and value are the three vectors on which self-attention model works. The vectors can be created by applying learned linear projections. The output of the BiLSTM is taken as an input to the self-attention model. To obtain the query, key, and value, three feedforward layers were used. For 1-layer stacked configuration the process needs no repetition. Attention width of 256 is used and Sigmoid is used as an activation function in the self-attention mechanism. By averaging out the last layer (output) of self-attention and the results passed through feed-forward layer, a self-attention map is created, which will be used for future classification. In the end, the image representative vector passes through the dense layer. *Tanh* is used by all feed-forward layers as an activation function and the final prediction is achieved using the SoftMax function.

All the experiments were executed on single GTX1080 Ti, with a 3.3 GHz and 32 Gb RAM and i7 processor. The model was implemented in KERAS 2.0.8 with Tensorflow 1.7 backend, using cuDNN 5.1.5 and CUDA 9.1.

**Results & Discussion**

The dataset is downloaded from https://idr.openmicroscopy.org/webclient/?experimenter=-1 [29], which consists of left ventricular heart tissue histopathology images obtained by carrying out EMB on patients, with end-stage heart failure who are clinically diagnosed as suffering from either as ischemic cardiomyopathy or non- ischemic cardiomyopathy. The dataset consists of both male and female patients. The average age of patients suffering from end-stage heart failure is $53.4 \pm 15.3$ years. The proposed self-attention based BiLSTM-CNN was used for an automated classification cardiomyopathy. In order to achieve so, we first differentiate the ischemic and non- ischemic cardiomyopathy images. The total size of the samples is 94, in which 51 samples were diagnosed as suffering from ischemic and 43 samples were determined to be suffering from non-ischemic cardiomyopathy. 65 images were used to train the networks, and 29 images were used to test the networks. The performance of the models was evaluated in terms of accuracy, sensitivity, and specificity.

Although, self- attention is used to generate attention aware features by using attention modules consisting of feed-forward structures. Attention aware features are directly dependent on depth (i.e. the depth of the sample used) which changes as depth increases. At each stage, the mean absolute response of the output layer is calculated for a better understanding of self-attention model. In the proposed model, we used attention one time for evaluation of images. The self-attention mechanism is used to store relevant information by suppressing the noise. It makes optimization much easier and also helps in classifying the represented features. This technique is applied to train the network required for assessment. Thus, a new classifier model was developed using a pre-trained network (both inception-V3 and LSTM) that worked more effectively in differentiating histopathological images based on deep learning algorithm self-attention mechanism. Figures 3 (a) and (b) show the performance curves of accuracy vs epoch and loss vs epoch, of the proposed model during for the training dataset respectively. By using adequate number of classified images during training the accuracy of the proposed model can reach 95.38 % for the training datasets while discriminating ischemic and non-ischemic images based on histopathology images.

To compare the performance of the proposed model two other models were also trained on the given datasets. An inception-V3 model, inception-V3 with LSTM and the proposed based on inception-V3 with BiLSTM and self-attention were trained and tested [30, 31]. Table 1 shows the comparative performance metrics of the different models on the training datasets. The proposed model has a better performance as compared to other conventional CNN models such as inception-V3 and inception-V3 with LSTM and in terms of accuracy, sensitivity and specificity to predict the cardiac outcome.

**Table 1**. Performance metrics of different models on the training datasets of H&E stained images.

| Model | Accuracy | Sensitivity | Specificity |
|---|---|---|---|
| Inception-V3 | 92.30 % | 93.75 % | 90.90 % |
| Inception-V3 + LSTM | 93.84 % | 93.93 % | 93.75 % |
| Proposed Model (Inception-V3 + BiLSTM + Self-Attention) | 95.38 % | 96.87 % | 93.93 % |

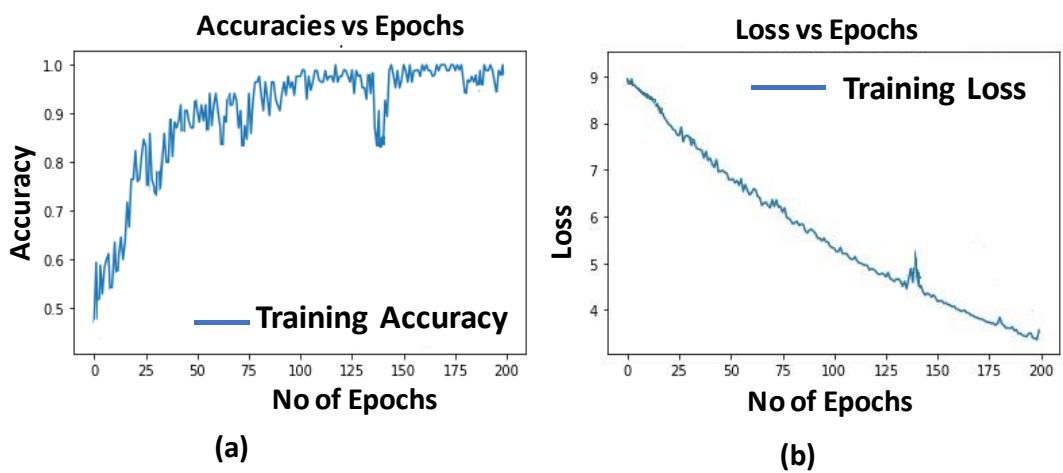

**Fig. 3.** (a, b) Performance curves of accuracy vs epochs and loss vs epochs for the training dataset, respectively of the proposed model (self-attention with BiLSTM-CNN).

The proposed model achieved 93.10 % accuracy, 93.33% sensitivity and 86.66 % specificity for the testing dataset. Our results reveal that we achieved a much better performance from the proposed model as compared to other two models based on CNN and CNN-LSTM on the small training dataset. Figure 4 shows histopathological images that were misclassified from all the models including the proposed one. Figure 4 (a, b) are misclassified as ischemic and non-ischemic cardiomyopathy, respectively. The primary reason for the network failure is that sometimes heart tissues have both ischemic and non-ischemic characteristic of cardiomyopathy, as a significant region of heart tissues is influenced by both coronary artery disease as well as some dysfunctional organs. This makes the classification between ischemic and non-ischemic cardiomyopathy hard for the network as well as pathologists.

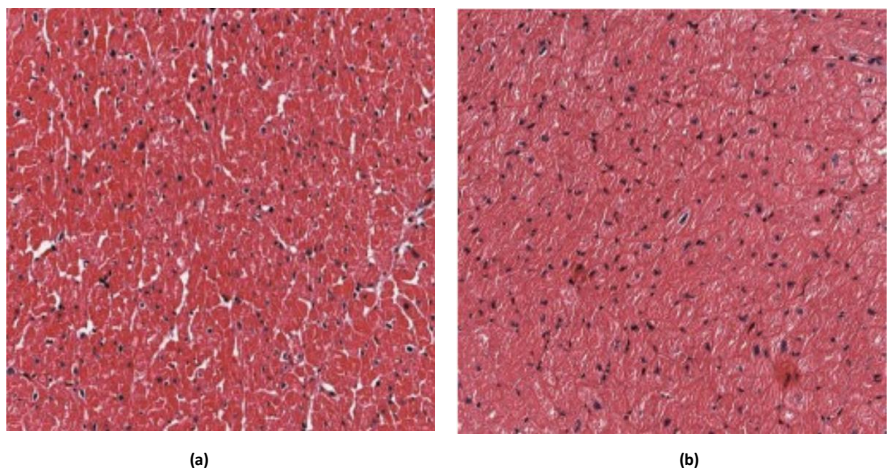

(a)          (b)

**Fig. 4.** Evaluation of misclassified histopathological images.

**Conclusion**

The proposed model demonstrates promising results and outperforms the state-of-the-art deep learning models for an automatic classification of ischemic and non-ischemic cardiomyopathy using histopathological images. The model is an integration of CNN (inception-V3) with BiLSTM and self-attention map; it improves the accuracy of classification even when trained on a small dataset. Data analysis on the testing dataset, using the proposed model revealed that it has attained a high accuracy of 93.10 %, sensitivity 93.33 % and specificity of 86.66 % in classifying ischemic and non-ischemic images. Some misclassification has also been noted. In the assessment of cardiac failure more study is needed to fully assess the potential of deep learning in clinical practice and its implications.

**Disclosure:** The author declares that there are no conflicts of interest.
**Acknowledgement:** Authors gratefully acknowledge the financial assistance from Science Engineering and Research Board, Government of India for the Project No. EMR/2016/000677.